\documentclass[letterpaper, 10 pt, conference]{ieeeconf}  

\IEEEoverridecommandlockouts                              
\usepackage{lmodern}
\usepackage{hyperref}
\usepackage{multirow}

\title{\LARGE \bf
Various Approaches to Aspect-based Sentiment Analysis
}

\author{ \parbox{3 in}{\centering Amlaan Bhoi\\
        Department of Computer Science\\
        University of Illinois at Chicago\\
        Chicago, IL, USA\\
        {\tt\small abhoi3@uic.edu}}
        \hspace*{ 0.5 in}
        \parbox{3 in}{ \centering Sandeep Joshi\\
        Department of Computer Science\\
        University of Illinois at Chicago\\
        Chicago, IL, USA\\
        {\tt\small sjoshi37@uic.edu}}
}

\begin{document}

\maketitle
\thispagestyle{empty}
\pagestyle{empty}

\begin{abstract}

The problem of aspect-based sentiment analysis deals with classifying sentiments (negative, neutral, positive) for a given aspect in a sentence. A traditional sentiment classification task involves treating the entire sentence as a text document and classifying sentiments based on all the words. Let us assume, we have a sentence such as "the \textit{acceleration} of this car is fast, but the \textit{reliability} is horrible". This can be a difficult sentence because it has two aspects with conflicting sentiments about the same entity. Considering machine learning techniques (or deep learning), how do we encode the information that we are interested in one aspect and its sentiment but not the other? Let us explore various pre-processing steps, features, and methods used to facilitate in solving this task.

\end{abstract}

\section{INTRODUCTION}

Aspect-based sentiment analysis is defined as given a text document $d_i$ and an aspect term/phrase $a_j$, we wish to determine the sentiment [-1, 0, +1] of the aspect term in the document [1]. In this setting, we assume aspect extraction has been performed and is not a primary task. Otherwise, we would also need to extract aspects and cluster them into some predefined class. 

We discuss the various pre-processing steps we used to generate features for our models. There is a plethora of approaches to aspect-based sentiment analysis. In this paper, we consider using classical machine learning techniques and two deep learning approaches. Finally, we look at the results, which model performed best and why that might be the case.

\section{TECHNIQUES}

We shall now explore the different techniques, methods, and features used in this experiment. We will divide the section into three sections: pre-processing steps and features common to all methods, classical machine learning models, and deep learning models. Some features and pre-processing steps may overlap between the two sections and that will be mentioned wherever appropriate.

\subsection{Common Pre-processing Steps \& Features}

We start with the first feature we use called \textbf{ID-encoding}. ID-encoding works by assigning each unique aspect in the text corpus (collection of documents/sentences) a unique ID that cannot be repeated. We then encode that id in a zero-vector based on the original aspect term's location. Formally, for each unique aspect term $a_i$ in the corpus, we assign a unique ID $id_i$ and store it in a dictionary as a key-value pair. Then, for each sentence $s_j$, we create a \textit{zero-vector} and replace zeros with $id_i$ in those aspect locations. We call this new vector an aspect-sequence. For example, if we have the sentence "the \textit{battery life} of the phone is too short", then assuming our unique ID is x, our aspect-sequence becomes:

\begin{center}[0 x x 0 0 0 0 0 0]\end{center}

Another variation of this would be to treat the whole aspect term as one token. Then, our aspect-sequence becomes:

\begin{center}[0 x 0 0 0 0 0 0]\end{center}

Deep learning models do not work well with sequences of different lengths. Thus, we applied \textbf{zero-padding} [2] to pad the sequences to a fixed length. Our next approach to feature engineering was simple and is called \textbf{bit-masking}. This approach follows the same concept as ID-encoding, but instead of applying a unique ID to each aspect, we encode a 1 to each aspect term location. Thus, the previous example's bit-masking vector would become:

\begin{center}[0 1 1 0 0 0 0 0 0]\end{center}

Our final effort to encode aspect information into our model is called \textbf{location-encoding}. In this approach, for each aspect term $a_i$ and sentence $s_j$, we encode the location of each \textit{context word} $c_k$ with respect to the aspect term in the sentence. Following our previous example, our \textit{location-sequence} becomes:

\begin{center}[1 1 2 3 4 5 6]\end{center}

where we do not include the aspect term location and our aspect term is considered one location regardless of its size (single word or phrase). For tokenization of text, we used Keras' \textbf{text\_to\_word\_sequence} method [3].

\subsection{Machine Learning Models}

In our machine learning models, the specific feature engineering we did was \textbf{TF-IDF vectorization}. This converts our text sentences into TF-IDF vectors and the whole corpus becomes a TF-IDF matrix. The tf-idf score formulation is as follows:

$$\textup{tfidf}_{t,d} = \textup{tf}_{t, d} \times \textup{idf}_{t}$$

Please refer to the book by Manning et. al [5] for full details. Let us look at the various machine learning models we used for this experiment. The features mentioned above are used in all the models. We shall just give a brief overview for each. Every model except XGBoost is from the Scikit-Learn library [4].

\begin{itemize}
    \item \textbf{Naive Bayes:} \textit{Naive Bayes} classifiers apply Bayes' theorem with the assumption of conditional independence between every pair of features. We chose this classifier because it has shown to perform surprisingly well on text problems.
    \item \textbf{Decision Tree:} \textit{Decision tree} is a non-parametric learning method that predicts the value of a target variable by learning decision rules. The reason why we chose this is because decision trees can learn inherent rules available in the dataset that are not available to the user.
    \item \textbf{Support Vector Machines (SVM):} \textit{Support vector machines} have shown to be extremely capable of handling aspect-based sentiment analysis in domains such as but not limited to customer reviews on laptops and restaurants [6]. Unsurprisingly, SVMs performed the best in the classical machine learning methods we tried. We shall see those results later.
    \item \textbf{Random Forest Classifier:} \textit{Random forest} classifier is a meta-estimator that fits decision trees on sub-samples of the input dataset. It can be regarded as an ensemble method. Fitting various decision trees on sub-samples of data can help prevent overfitting. We chose this because we believed it can be a better classifier than vanilla decision trees.
    \item \textbf{Extra Trees Classifier:} \textit{Extra trees} classifier is another meta-estimator (also categorized as an ensemble method) which fits randomized decision trees (a.k.a extra trees) on various sub-samples of the dataset. The motivation behind using this method is the same as using random forest classifiers.
    \item \textbf{Extreme Gradient Boosting (XGBoost):} \textit{Extreme gradient boosting} is an optimized, parallel implementation of gradient boosting [7]. This algorithm is based on boosting of decision trees. Please refer to the paper by Friedman [7] for a better intuition behind XGBoost. We chose this algorithm because it proved to perform well on our previous machine learning experiments.
\end{itemize}

\subsection{Deep Learning Models}

The only specific pre-processing step we did for the deep-learning model is \textbf{stop-word removal} [8]. For word embeddings, we used Stanford's \textbf{GLoVE embeddings} [9]. Let us now explore the two deep learning approaches we tried. We shall also explore why one worked at another did not. Both methods are built upon on the LSTM [11] and Attention [12] concepts.

\begin{itemize}
    \item \textbf{Attention LSTM-RNN:} The first model we tried was vanilla \textit{LSTM-RNN} [10] with an attention layer to learn the weights of context words with respect to the aspect. This approach did not work well as a single attention layer is not sufficient to learn the abstract features between the aspect and context terms. The problem was also compounded by the fact that our encoding of features was not sufficient for a single layer pass to work. We tried Bidirectional LSTMs, dropout, recurrent dropout, MaskedGlobalAveragePooling, BatchNormlization, LeakyReLU, and more. The basic architecture that we built can be found \href{https://drive.google.com/file/d/1o3JxbKSCG8dg83EO5cMS0NBTEI2JuRtu/view}{here}.
    \item \textbf{Deep Memory Network:} The second model we implemented is the \textit{deep memory network} (a.k.a. MemNet) [12]. Instead of using just one attention+linear layer, it uses multiple layers called hops. Hops are needed for this task because the task requires multiple levels of abstraction. Each hop performs the aforementioned operation and feeds it into a softmax layer. The model calculates context attention as well as location attention. The input to this model is the input sentence tokenized with unique word IDs, aspect term/phrase, location of the aspect term, and a location vector denoting each context word's location with respect to the aspect term. This model performed the best as we shall see.
\end{itemize}

Attention LSTM-RNN is built using \textbf{Keras} [3] while the MemNet was built using \textbf{Tensorflow} [2]. All models were trained on Stratified K-Fold validation to ensure no bias and confirm the metrics of precision, recall, F-1 score, and overall accuracy for each dataset. For prediction, we shuffled and split the train and test dataset, saved the trained model, and loaded it back up for testing.

\section{RESULTS}

The dataset used for our experiments is a modified version of the SemEval 2016: Task 4 challenge [15]. We apply the algorithms on the Tech Reviews and Food Reviews domains.

In our experiments, \textbf{MemNet} worked best with an overall accuracy of 0.713 and 0.7866 on tech reviews and food reviews dataset respectively. We trained all models on Google Cloud Platform with 16 Intel Skylake CPUs, 60GB RAM, and Nvidia P100 GPU. The second best consistent performer was SVM with one-hot encoding the text. One surprisingly good model is the ETC on Tech dataset. However, the model failed to perform well on Food dataset with positive class F1-score at 0.3745 which is below par.


\begin{table*}[]
\centering
\caption{Test Results}
\label{my-label}
\begin{tabular}{|c|c|c|c|c|c|c|c|c|c|c|c|}
\hline
\multicolumn{12}{|c|}{TEST}                                                                                                                     \\ \hline
                                    &         & \multicolumn{3}{c|}{Positive Class} & \multicolumn{3}{c|}{Negative Class} & \multicolumn{3}{c|}{Neutral Class} &          \\ \hline
Classifier                          & Dataset & P         & R       & F1      & P        & R        & F1      & P        & R       & F1      & Accuracy \\ \hline
\multirow{2}{*}{Naive Bayes + OH}   & Tech    & 0.6349    & 0.6153  & 0.625   & 0.5722   & 0.5625   & 0.5673  & 0.3142   & 0.3586  & 0.335   & 0.5399   \\ \cline{2-12} 
                                    & Food    & 0.71466   & 0.6261  & 0.6674  & 0.3915   & 0.4037   & 0.3974  & 0.2458   & 0.3333  & 0.2829  & 0.5201   \\ \hline
\multirow{2}{*}{Decision Tree + OH} & Tech    & 0.6489    & 0.6256  & 0.6370  & 0.5073   & 0.5852   & 0.5435  & 0.3513   & 0.2826  & 0.3132  & 0.5397   \\ \cline{2-12} 
                                    & Food    & 0.7058    & 0.8130  & 0.7557  & 0.5405   & 0.3726   & 0.4411  & 0.4196   & 0.3560  & 0.3852  & 0.6269   \\ \hline
\multirow{2}{*}{SVM + OH}           & Tech    & 0.6945    & 0.7230  & 0.7085  & 0.6235   & 0.6306   & 0.6271  & 0.3780   & 0.3369  & 0.3563  & 0.6112   \\ \cline{2-12} 
                                    & Food    & 0.7231    & 0.8668  & 0.7885  & 0.5750   & 0.4285   & 0.4911  & 0.4318   & 0.2878  & 0.3454  & 0.6629   \\ \hline
\multirow{2}{*}{RFC + LE}           & Tech    & 0.5689    & 0.7274  & 0.6376  & 0.7079   & 0.6897   & 0.6975  & 0.5454   & 0.2916  & 0.3753  & 0.6247   \\ \cline{2-12} 
                                    & Food    & 0.4136    & 0.1783  & 0.2438  & 0.6579   & 0.9040   & 0.7599  & 0.3448   & 0.1514  & 0.2061  & 0.6147   \\ \hline
\multirow{2}{*}{XGBoost + LE}       & Tech    & 0.5669    & 0.8435  & 0.6771  & 0.7768   & 0.6976   & 0.7342  & 0.6220   & 0.1911  & 0.2882  & 0.6511   \\ \cline{2-12} 
                                    & Food    & 0.6387    & 0.1151  & 0.1877  & 0.6436   & 0.9655   & 0.7713  & 0.5018   & 0.1167  & 0.1779  & 0.6333   \\ \hline
\multirow{2}{*}{ETC + LE}           & Tech    & 0.6414    & 0.8121  & 0.7159  & 0.7973   & 0.7440   & 0.7687  & 0.6275   & 0.4043  & 0.4873  & 0.7021   \\ \cline{2-12} 
                                    & Food    & 0.5027    & 0.3002  & 0.3745  & 0.6907   & 0.8790   & 0.7728  & 0.3732   & 0.2000  & 0.2550  & 0.6363   \\ \hline
\multirow{2}{*}{MemNet}             & Tech    & 0.8249    & 0.7694  & 0.7960  & 0.7008   & 0.6776   & 0.6835  & 0.4475   & 0.5178  & 0.4669  & \textbf{0.7130}   \\ \cline{2-12} 
                                    & Food    & 0.8446    & 0.9241  & 0.8819  & 0.7222   & 0.6921   & 0.7044  & 0.6175   & 0.4401  & 0.5051  & \textbf{0.7866}   \\ \hline
\end{tabular}
\end{table*}

\section{CONCLUSION}

In conclusion, it is apparent that MemNet works best on this task based on our experiments. Surprisingly, ETC performed better than expected on the tech reviews dataset. The main problem is still the class imbalance. This leads to low precision, recall, and F-1 scores for the neutral class. This can further be fixed by oversampling or resampling. More aspects we can tackle are dependency parsing as a feature, opinion dictionary, sentiment dictionary, using ELMo embeddings instead of GLoVe embeddings, modularizing our pipeline for faster experiments, and gaining a deeper understanding of why some models do not work on these datasets or these types of problems. For other future experiments, we want to try Tree-LSTMs [13] and Interactive Attention Networks [14].

\addtolength{\textheight}{-12cm}   









\begin{thebibliography}{99}

\bibitem{c1} Liu, Bing. \textit{Web data mining: exploring hyperlinks, contents, and usage data}. Springer Science \& Business Media, 2007.
\bibitem{c2} Abadi, Martin, et al. "TensorFlow: A System for Large-Scale Machine Learning." \textit{OSDI}. Vol. 16. 2016.
\bibitem{c3} Chollet, Francois. "Keras." (2015): 128.
\bibitem{c4} Buitinck, Lars, et al. "API design for machine learning software: experiences from the scikit-learn project." \textit{arXiv preprint arXiv:1309.0238} (2013).
\bibitem{c5} Schutze, Hinrich, Christopher D. Manning, and Prabhakar Raghavan. \textit{Introduction to information retrieval}. Vol. 39. Cambridge University Press, 2008.
\bibitem{c6} Kiritchenko, Svetlana, et al. "NRC-Canada-2014: Detecting aspects and sentiment in customer reviews." Proceedings of the 8th International Workshop on Semantic Evaluation (SemEval 2014). 2014.
\bibitem{c7} Friedman, Jerome H. "Greedy function approximation: a gradient boosting machine." Annals of statistics (2001): 1189-1232.
\bibitem{c8} Bird, Steven, Ewan Klein, and Edward Loper. Natural language processing with Python: analyzing text with the natural language toolkit. " O'Reilly Media, Inc.", 2009.
\bibitem{c9} Pennington, Jeffrey, Richard Socher, and Christopher Manning. "Glove: Global vectors for word representation." Proceedings of the 2014 conference on empirical methods in natural language processing (EMNLP). 2014.
\bibitem{c10} Hochreiter, Sepp, and Jurgen Schmidhuber. "Long short-term memory." \textit{Neural computation} 9.8 (1997): 1735-1780.
\bibitem{c11} Luong, Minh-Thang, Hieu Pham, and Christopher D. Manning. "Effective approaches to attention-based neural machine translation." \textit{arXiv preprint arXiv:1508.04025} (2015).
\bibitem{c12} Tang, Duyu, Bing Qin, and Ting Liu. "Aspect level sentiment classification with deep memory network." \textit{arXiv preprint arXiv:1605.08900} (2016).
\bibitem{c13} Tai, Kai Sheng, Richard Socher, and Christopher D. Manning. "Improved semantic representations from tree-structured long short-term memory networks." \textit{arXiv preprint arXiv:1503.00075} (2015).
\bibitem{c14} Ma, Dehong, et al. "Interactive Attention Networks for Aspect-Level Sentiment Classification." arXiv preprint arXiv:1709.00893 (2017).
\bibitem{c15} Pontiki, Maria, et al. "SemEval-2016 task 5: Aspect based sentiment analysis." \textit{Proceedings of the 10th international workshop on semantic evaluation (SemEval-2016)}. 2016.

\end{thebibliography}
\end{document}